\pdfoutput=1

\documentclass[11pt]{article}

\usepackage[]{acl}

\usepackage{times}
\usepackage{latexsym}

\usepackage[T1]{fontenc}

\usepackage[utf8]{inputenc}

\usepackage{microtype}

\usepackage{booktabs,caption}
\usepackage[flushleft]{threeparttable} 
\usepackage{lineno}
\usepackage[pdftex]{graphicx} 
\usepackage{amsmath}
\usepackage{makecell}
\usepackage{float}
\usepackage{multicol,babel,float,blindtext}
\usepackage{amssymb}
\usepackage{multirow}
\newcommand{\Ours}[0]{\textsc{WeaklY superVised genERative documeNt parser}}
\newcommand{\ours}[0]{\textsc{Wyvern}}
\newcommand{\vone}[0]{\textsc{POT}}

\newcommand{\nj}[0]{\textsc{nj}}

\newcommand{\njw}[0]{\textsc{nj-w}}
\newcommand{\rk}[0]{\textsc{rk}}

\newcommand{\rj}[0]{\textsc{rj}}

\newcommand{\ted}[0]{n\textsc{TED}}

\title{Cost-effective End-to-end Information Extraction for\\ Semi-structured Document Images} 

\author{Wonseok Hwang$^{a,}$\thanks{\quad Most work done while these authors were at NAVER.} \quad Hyunji Lee$^{b,*}$ \quad Jinyeong Yim$^c$ \quad Geewook Kim$^c$ \quad Minjoon Seo$^{b,*}$\\
  $^a$LBox \quad $^b$KAIST \quad $^c$Clova, NAVER\\
  \texttt{wonseok.hwang@lbox.kr} \quad \texttt{hyunji.amy.lee@kaist.ac.kr}
  \\\texttt{\{jinyeong.yim,gw.kim\}}\texttt{@navercorp.com}
  \\\texttt{minjoon@kaist.ac.kr}\\
}

\begin{document}
\maketitle

\begin{abstract}

A real-world information extraction (IE) system for semi-structured document images often involves a long pipeline of multiple modules, whose complexity dramatically increases its development and maintenance cost.
One can instead consider an end-to-end model that directly maps the input to the target output and simplify the entire process.
However, such generation approach is known to lead to unstable performance if not designed carefully.
Here we present our recent effort on transitioning from our existing pipeline-based IE system to an end-to-end system focusing on practical challenges that are associated with replacing and deploying the system in real, large-scale production. By carefully formulating document IE as a sequence generation task, we show that a single end-to-end IE system can be built and still achieve competent performance.

\end{abstract}

\section{Introduction} \label{sec: intro}
\begin{figure}[t]
\centering
\includegraphics[width=0.44\textwidth]{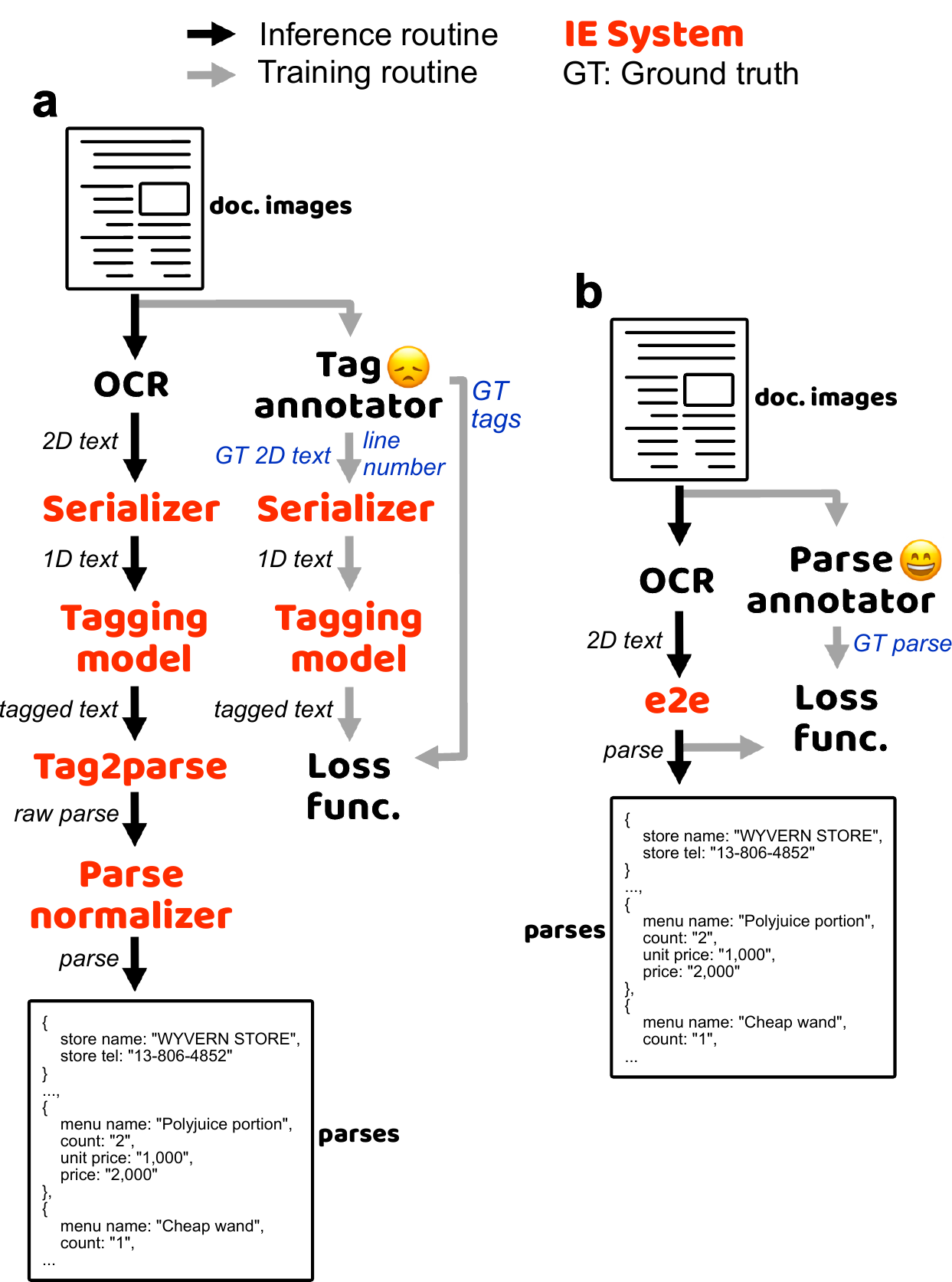}
\caption{The scheme of (a) our tagging based IE system and (b) the end-to-end IE system proposed in this study. } 
\label{fig_v1_vs_v3task}
\vspace{-4mm}

\end{figure}

Information extraction (IE) for semi-structured documents is an important first step towards automated document processing.
One way of building such IE system is to develop multiple separate modules specialized in each sub-task. For example, our currently-deployed IE system for name card and receipt images, \vone\ \citep{hwang2019pot},\footnote{In October 2020, the system receives approximately 350k name cards and 650k receipts queries per day.} consists of three manually engineered modules and one data-driven module (Fig.~\ref{fig_v1_vs_v3task}a). This system first accepts text segments and their 2D coordinates (we dub it ``2D text'') from an OCR system and generates pseudo-1D-text using a serializer. The text is then IOB-tagged and mapped to a raw, structured parse. Finally, the raw parse is normalized by trimming and reformatting with regular expressions.

Although \vone\ has shown satisfactory performance to its customers, its long pipeline increases the development 
cost: (1) each module should be engineered and fine-tuned by experts per domain, and (2) each text segment should be annotated manually for the training. 
As entire structural information of documents should be recovered from a set of individual tags, the difficulty of the annotation process increases rapidly with the layout complexity.

One can consider replacing the entire pipeline into a single end-to-end model but such model is known to show relatively low accuracy in structured prediction tasks without careful modeling \citep{dyer2016rnng,dongC2F2018,yin2018TRANX,fernandez-astudillo2020transition_transformer}.
Here, we present our recent effort on such transition to replace our product \vone\ with an end-to-end model.
In the new IE system, which we call \ours\footnote{\tiny{\Ours}}, we formulate the IE task as a sequence generation task equipped with a tree-generating transition module.
By directly generating a sequence of actions that has a one-to-one correspondence with the final output (parse), the system can be trained without annotating intermediate text segments or developing and maintaining four different IE modules in \vone\  (orange texts in Fig.~\ref{fig_v1_vs_v3task}a).
This also leads to the dramatic reduction in cost (Tbl.~\ref{tbl_cost}). 

To achieve service-quality accuracy and stability with a sequence generation model while minimizing domain-specific engineering, we start from a 2D Transformer baseline and experiment with the following components: (1) copying mechanism, (2) transition module, and (3) employing preexisting weak-label data. We observe that \ours\ achieves comparable performance with \vone\ when it is trained with a similar amount of, yet low-cost data, and achieves higher accuracy by leveraging a large amount of preexisting weak supervision data which cannot be utilized by the pipeline-based system. This signifies that turning a long machine learning pipeline into an end-to-end model is worth considering even in real production environment.




\paragraph{Related Work}
Most of previous works formulate semi-structured document IE task as text tagging problem \citep{palm2017cloundscan,katti2018chargrid,zhao2019cutie,xu2019_layoutLM,denk2019arXiv,majumder2020representation,liu2019graph,yu2020_pick,qian2019graphie,hwang2019pot}.
\citealp{hwang2020spade} formulates the task as spatial dependency parsing which is essentially an another tagging approach where the inter-text segment relations are tagged.
Although all previous studies have shown the competency of the proposed methods on their own tasks, they have following limitation: individual text segments should be labeled by appropriate tags for the model training. Since the tags require domain and layout-dependent modification for each task, the appropriate annotation tools should be developed together. 
The difficulty of annotation rapidly increases when documents show multiple information hierarchy necessitating grouping between fields (for example, see \texttt{name}, \texttt{price} fields in Fig.~\ref{fig_tree}a).
  On the other hand, we formulate the IE task as a sequence generation problem where only parses are required for the training.

\section{Model}\label{sec: model}
\ours\ consists of the following three major modules: (1) the Transformer encoder that accepts 2D text, (2) the decoder for sequence generation, and (3) the transition module that converts the generated sequence into parse tree.

\paragraph{2D Transformer Encoder}
We use Transformer \citep{vaswani2017transformer} with the following modification for encoding 2D text \citep{hwang2020spade}.
The input vectors are generated using the following five features: token, x-coordinate, y-coordinate, character height, and text orientation. Like BERT, each feature is represented as integers and maps to a trainable embedding vector. The each coordinate is quantized into 120 integers, character height into 6, and text orientation into 2 integers. 
The resulting embeddings are summed into a single input vector. 
Unlike original transformer, the position embedding for word ordering is omitted.

\paragraph{Decoder}
We use Transformer decoder equipped with the gated copying mechanism \citep{gu2016copynet,see2017gatedcopy}.
At each time $t$, the probability of copying individual input tokens is calculated via inner product between the contextualized inputs from the last Transformer layers of the encoder ($\{h_e\}$) and the decoder ($h_d(t)$).
The resulting probability is added to the generation probability of corresponding tokens gated by the probability $p_{\text{gate}}$.
$p_{\text{gate}}$ is calculated by linearly projecting the concatenated vector of the $h_d(t)$ and the sum of $\{h_e\}$ each weighted by $h_d(t)$.

\begin{figure}[t!]
\centering
\includegraphics[width=0.475\textwidth]{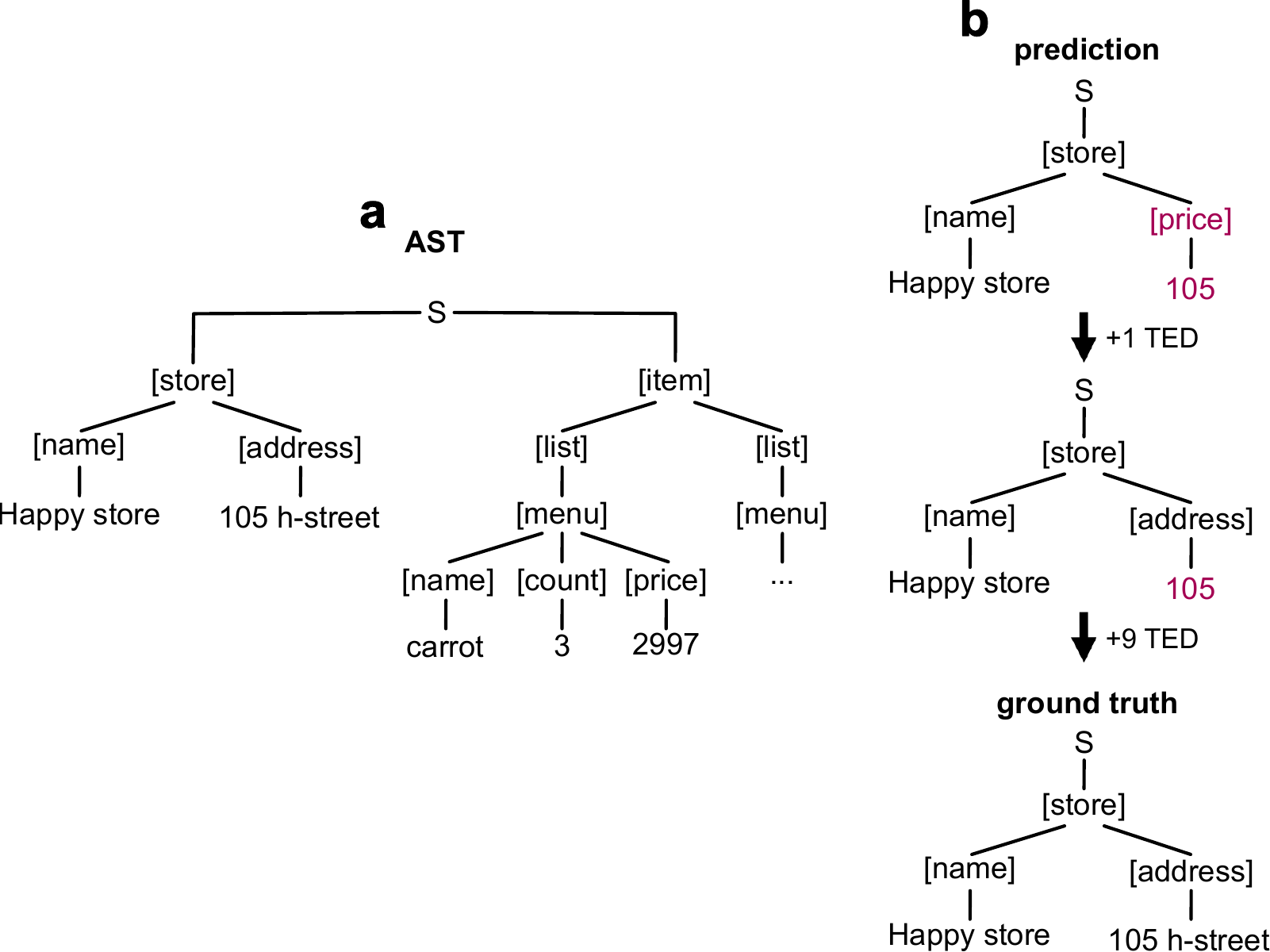}
\caption{Tree representation of document. (a) An example of abstract syntax tree (AST). 
(b) An example of TED calculation. 
}
\label{fig_tree}
\vspace{-4mm}
\end{figure}

\paragraph{Transition module}

All parses are uniformly formatted following JSON.
However, direct generation of JSON strings requires unnecessary long steps because (1) all syntactic tokens (\texttt{\{},\texttt{:}, \texttt{\}}) are generated separately, and (2) ``key'' often consists of multiple tokens.
To minimize the generation length while constraining the search space, we propose to convert JSON-formatted parses into corresponding abstract syntax trees (ASTs) (Fig.~\ref{fig_tree}a).
Under this formulation, the sequence of generated tokens is interpreted as a sequence of AST generating actions.
The actions are later converted into a JSON-formatted parse using the push-down automaton.
Our transition module consists of three types of actions \texttt{NT(key)},  \texttt{GEN(token)}, and \texttt{REDUCE} based on previous studies \citep{yin2018TRANX,dyer2016rnng}. 
\texttt{NT(key)} generates non-terminal node representing \texttt{key} of JSON. 
The special tokens representing the type of individual fields are interpreted as the corresponding actions. 
\texttt{GEN(token)} generates corresponding \texttt{token}. 
\texttt{REDUCE} indicates the completion of generation in a current level and the systems moves to a higher level.
The process is demonstrated with an example in Tbl.~\ref{tbl_parse_gen}.
A sequence of actions can be uniquely determined for a given AST by traveling the tree in depth-first, and left-right order.

\begingroup
\setlength{\tabcolsep}{1.4pt} 
\renewcommand{\arraystretch}{0.6} 
\begin{table}[t!]
\captionsetup{font=small} 
  \caption{An example of parse generation process. The corresponding parse tree is shown in Fig.~\ref{fig_tree}a.
  }
  \label{tbl_parse_gen}
  \tiny
  \centering
  \begin{tabular}{lccl}
    \toprule
    $t$ & Output token & Action & Parse  \\
    \midrule
    0 & - & - & \{S:\\
    1 & [store] & \texttt{NT(store)} &\{S:\{store:  \\
    3 & [name] & \texttt{NT(name)} &\{S:\{store: \{name: \\
    4 & Happy & \texttt{GEN(Happy)} &\{S:\{store: \{name: Happy \\
    5 & store & \texttt{GEN(store)} &\{S:\{store: \{name: Happy store \\
    6 & [reduce]& \texttt{REDUCE} &\{S:\{store: \{name: Happy store,\\
    7 & [address]& \texttt{NT(address)} &\{S:\{store: \{name: Happy store, address: \\
    ...& & & \\
    \bottomrule
  \end{tabular}
  \vspace{-4mm}

\end{table}
\endgroup

\section{Datasets and Setups}

\paragraph{Datasets}
 Due to confidential issues related to industrial documents, we use our four large-scale internal datasets: three strong-label (\nj, \rk, \rj) and one weak-label (\njw) datasets.
The properties are summarized in Tbl.~\ref{tbl_datasets}. 
\begingroup
\setlength{\tabcolsep}{2pt} 
\renewcommand{\arraystretch}{0.8} 
\begin{table}[h!]
\centering
\begin{threeparttable}

\captionsetup{font=small} 
  \caption{
  The dataset properties.
  }
  \label{tbl_datasets}
  \tiny
  \centering
  \begin{tabular}{l@{\extracolsep{\fill}}cccccc}

    \toprule
     Name & Task & Lang &\makecell{\# of field\\keys} & Tree depth& \makecell{\# of examples\\(train:dev:test)} & \makecell{Input token\\length (test)}  \\
     \midrule
     \midrule
     \nj& name card & JPN & 12 & 3& 22k:256:256 & $115\pm36$    \\ 
     \rk & receipt & KOR & 21 &3  & 44k:600:781 & $323\pm95$   \\ 
     \rj & receipt & JPN & 39 & 3 & 62k:600:3,330& $165\pm122$   \\ 
     \midrule
     \njw & name card & JPN & 13 & 2 & 6m:1,020:1,020&  $124 \pm 38$ \\ 
    \bottomrule
  \end{tabular}
\end{threeparttable}
\vspace{-4mm}
\end{table}
\endgroup

\paragraph{Baseline}
We compare \ours\ to \vone\ that consists of four separated modules (Fig.~\ref{fig_v1_vs_v3task}a, Sec.~\ref{sec: intro}). The technical details are described in Sec.~\ref{sec: pot detail}.

\paragraph{Evaluation}
The parses consist of hierarchically grouped key-value pairs, where a key indicates a field type such as \texttt{menu}, and a value is either a corresponding text sequence or a sub parse forming a recursive structure (e.g., \texttt{menu} under \texttt{item} in Fig.~\ref{fig_tree}a).
We conduct extensive evaluation with three different metrics; $F_1$, \ted, and A/B-test.
$F_1$ is calculated by counting the number of exactly matched key-value pairs. Since $F_1$ ignores partially correct prediction (even a single character difference is counted as wrong), we use another metric \ted, a tree edit distance normalized by the number of nodes.
\ted\ considers both lexical and structural differences between parse trees (Fig.~\ref{fig_tree}b).
Finally, we conduct A/B-test for full evaluation of the models by human. See Sec.~\ref{sec: evaluation detail} for the technical details of evaluation methods.

\paragraph{Experimental setup}
The model is initialized with pretrained multi-lingual BERT \citep{devlinBERT2018,wolf2020huggin} and trained by Adam optimizer \citep{kingam2015adam} with learning rate 2e-5 or 3e-5. The decay rates are set to $\beta_1 = 0.9, \beta_2 = 0.999$. After the initial rapid learning phase, which typically takes 1--2 weeks on 2--8 NVIDIA P40 gpus, the learning rate is set to 1e-5 for the stability and the training continues up to one month. The batch size set to 16--32. During inference, beam search is employed.
In receipt IE task, the training examples are randomly sampled from three tasks--Japanese name card, Korean receipt, and Japanese receipt--while sharing the weights. We found this multi-domain setting leads to the faster convergence.

\section{Results}
   
\begingroup
\setlength{\tabcolsep}{6pt} 
\renewcommand{\arraystretch}{1} 
\begin{table}[b!]
\centering
\begin{threeparttable}
\captionsetup{font=small} 
  \caption{Performance table of name card IE task.}
  \label{tbl_results_nj}
  \tiny
  \centering
  \begin{tabular}{lccc}
  \toprule
   & is E2E  & $1-F_1$ (\%) & \ted\ (\%)\\
    \midrule
    \vone~\citep{hwang2019pot}  &  & \textbf{10.5} & 8.22 \\
    2D Transformer$^\dag$~\citep{vaswani2017transformer} & \checkmark & 19.7 & 11.7 \\ 
    \ours~(\textbf{Proposed}) & \checkmark & 11.0 & \textbf{7.08}   \\ 
    \midrule
    \ours~w/ weak-label pretraining & \checkmark   & \textbf{8.1} & \textbf{5.93} \\ 
    \bottomrule
  \end{tabular}
 \begin{tablenotes}
 \item[$\dag$]{Transformer model with minimal modification to encode 2D texts (Sec.~\ref{sec: model}).}
 \end{tablenotes}

 \vspace{-4mm}
\end{threeparttable}
\end{table}
\endgroup

 \paragraph{\ours\ shows competent performance} \label{sec: wyvern}
 We first validate \ours\ on Japanese name card IE task (\nj). The comparable scores of \ours\ with respect to \vone\ shows the effectiveness of our end-to-end approach (Tbl.~\ref{tbl_results_nj}, 1st row vs 3rd row).
 Note that a naive application of Transformer encoder-decoder model shows dramatic performance degradation (2nd row) highlighting the importance of the careful modeling.
 Here, $F_1$ score is based on exact match and \ours\ \textbf{generates} accurate parses even for non semantic text like telephone numbers (Tbl.~\ref{tbl_individual_fields_nj} last three columns in Appendix). 
 Higher precision can be achieved by controlling the recall rate (Fig.~\ref{fig_pr_curves} in Appendix).
 Our generative approach also enables automatic text normalization and OCR error correction.
 Without the transition module, $F_1$ error increases by 1.3\%.
 
 \paragraph{Utilizing large weak-label data significantly enhances accuracy} \label{sec: weak-label}
Often times, a weak-label data (document and parse pairs) already exists because it is what gets stored in databases. Especially, when a human-driven data collection pipeline has been existed, the amount of accumulated data can be huge.
To leverage the preexisting weak-label data in the database, 
we first train \ours\ using \njw\ that consists of 6m weakly labeled Japanese name cards and fine-tune it using \nj. The fine-tuning step is required as parses in \njw\ and \nj\ have multiple distinct properties (Sec.~\ref{sec: nj vs njw}).
The results shows \ours\ outperforms \vone\ by -2.4 $F_1$ error and -2.29 \ted\ (Tbl.~\ref{tbl_results_nj}, bottom row).
Note that \vone\ requires strong-label data that should be annotated from scratch unlike \ours.

\paragraph{\ours\ can generate complex parse trees} 
We further validate \ours\ on Korean (\rk) and Japanese (\rj) receipt IE tasks where parses have more complex structure (Fig.~\ref{fig_tree}, Tbl.~\ref{tbl_datasets}). 
\ours\ shows higher performance compared to \vone\ in \rk\ (Tbl.~\ref{tbl_results_receipt}, 1st row vs 3rd row) even for numeric fields (Tbl.~\ref{tbl_individual_fields_rk} last ten columns in Appendix). However in \rj, it shows the lower performance (2nd row vs 4th row). 
The lower performance in \rj\ may be attributed to its complex parse tree that consists of a total of 39 fields (Tbl.~\ref{tbl_datasets}).
 

\begingroup
\setlength{\tabcolsep}{9pt} 
\renewcommand{\arraystretch}{1} 
\begin{table}[t!]
\centering
\begin{threeparttable}
\captionsetup{font=small} 
  \caption{Performance table of receipt IE task.}
  \label{tbl_results_receipt}
  \tiny
  \centering
  \begin{tabular}{lccc}
  \toprule
   & Lang.  & $1-F_1$ (\%) & \ted\ (\%)\\
    \midrule
    \multirow{ 2}{*}{\vone}  & KOR & 15.7 & 12.1 \\ 
                                                  & JPN & \textbf{20.5} & \textbf{17.6} \\
 
    \multirow{ 2}{*}{\ours}   & KOR & \textbf{13.0} &  \textbf{9.8} \\
                                                  & JPN & 25.8 & 21.1 \\
                                                  
    \bottomrule
  \end{tabular}
 
 \vspace{-4mm}
\end{threeparttable}
\end{table}
\endgroup

\paragraph{\ours\ is preferred in A/B-test}
We conduct three A/B-tests between \vone\ (\textsc{P}) and \ours\ (\textsc{W}) on Japanese name card and Korean receipt IE tasks with varying training set.
\ours\ achieves comparable performance with \vone\ (Tbl.~\ref{tbl_ab}, 1st panel, the neutral rate $\sim$ 50\%) and better performance with the use of preexisting weak-label data (final row). When name cards have complex layouts, \ours\ is always favoured (Sec.~\ref{sec: ab detail}). 

\begingroup
\setlength{\tabcolsep}{5pt} 
\renewcommand{\arraystretch}{0.8} 
\begin{table}[t!]
\captionsetup{font=small} 
  \caption{A/B-test for human evaluation. 
  }
  \label{tbl_ab}
  \tiny
  \centering
  \begin{tabular}{ccccccc}
      \toprule
    \multicolumn{2}{c}{} &                  
    \multicolumn{2}{c}{Training set} &                  
    \multicolumn{3}{c}{Survey results} 
    \\
    \cmidrule(r){3-4}
    \cmidrule(r){5-7}

    Task &  \# of samples &\textsc{P} &  \textsc{W} & Neutral    &\textsc{P} is better  & \textsc{W} is better  \\
    \midrule
    \nj &240 &\nj & \nj& 45.4 & \textbf{28.8} & 25.8 \\
    \rk & 240 &\rk & \nj, \rk, \rj & 47.9 & \textbf{26.3} & 25.8 \\
    \midrule
    \nj &80 &\nj & \nj, \njw& 45.0 & 23.8 & \textbf{31.2} \\
    \bottomrule
  \end{tabular}
  \vspace{-4mm}

\end{table}
\endgroup


\paragraph{\ours\ is cost-effective} \label{sec: cost}
Training \vone\ requires a strong-label (tagging of individual text segments) data. The tags should convey information of field type, intra-field grouping (collecting text segments belong to the same field), and inter-field grouping (e.g. \texttt{name}, \texttt{count}, and \texttt{price} in Fig.~\ref{fig_tree}b form a group).
On the other hand, \ours\ is trained by using a weak-label data, i.e. parses (structured text) which can be conceived more easily and typed directly by annotators. 
This fundamental difference in the labels bring various advantages to \ours\ (\textsc{W}) compared to \vone\ (\textsc{P}).
Here we focus on the cost and perform semi-quantitative analysis based on our own experiences.
We split the cost into five categories: annotation cost (Annot.) for preparing training dataset, communication cost (Comm.) for teaching data annotators, annotation tool development cost (Tool dev.), maintenance cost (Maint.) that involves data denoising \& post collection, and inference cost related to serving. 
The result is summarized in Tbl.~\ref{tbl_cost}. The details of the cost estimation process is presented in \ref{sec: cost detail}.
 
\begingroup
\setlength{\tabcolsep}{2pt} 
\renewcommand{\arraystretch}{1} 
\begin{table}[th!]
\centering
\begin{threeparttable}

\captionsetup{font=small} 
  \caption{Cost analysis table. 
  }
  \label{tbl_cost}
  \tiny
  \centering
  \begin{tabular}{ccccccc}
    \toprule
    Model &Data & Annot.$^\dagger$ & Comms. & Tool dev. &  Maint. & \makecell{Inference\\time (s)}\\
    \midrule
    
    \textsc{P}
    & strong-label
    & $\sim 12$
    & $\sim$ 1--2 PM  
    & $\sim$ 0.3 PM
    & Need expert 
    & \nj: 0.4, \rk: 1.6
    \\
    \textsc{W}
    & weak-label
    & $\sim60$
    & $\sim$0
    & $\sim$0
    & No expert 
    & \nj: 1.4, \rk: 2.3
    \\
    \bottomrule
  \end{tabular}
\begin{tablenotes}
\tiny
\item[$\dagger$]{\texttt{\# of documents/person$\cdot$hr}}
\end{tablenotes}
\end{threeparttable}
\vspace{-4mm}
\end{table}
\endgroup

\section{Conclusion}
Here, we present \ours\ that is trained via weak supervision. \ours\ achieves competent performance compared to the strongly supervised model while its development and maintenance cost is significantly lower. 
We also provide cost analysis of developing IE systems for semi-structured documents. 
Currently, \ours\ shows slower training and inference time compared to the tagging based approach like other autoregressive models. Future work will be focused on optimizing the training protocol and the inference speed.

\section*{Acknowledgements}
We thank Kyunghyun Cho for the helpful comment about the transition module, and Sohee Yang for the critical comment on the figures.

\bibliography{post_ocr_parsing3}

\begin{thebibliography}{23}
\expandafter\ifx\csname natexlab\endcsname\relax\def\natexlab#1{#1}\fi

\bibitem[{{Denk} and {Reisswig}(2019)}]{denk2019arXiv}
Timo~I. {Denk} and Christian {Reisswig}. 2019.
\newblock \href {http://arxiv.org/abs/1909.04948} {{BERTgrid: Contextualized
  Embedding for 2D Document Representation and Understanding}}.
\newblock \emph{arXiv e-prints}, page arXiv:1909.04948.

\bibitem[{Devlin et~al.(2018)Devlin, Chang, Lee, and
  Toutanova}]{devlinBERT2018}
Jacob Devlin, Ming{-}Wei Chang, Kenton Lee, and Kristina Toutanova. 2018.
\newblock {BERT:} pre-training of deep bidirectional transformers for language
  understanding.
\newblock \emph{NAACL}.

\bibitem[{Dong and Lapata(2018)}]{dongC2F2018}
Li~Dong and Mirella Lapata. 2018.
\newblock \href {https://www.aclweb.org/anthology/P18-1068} {Coarse-to-fine
  decoding for neural semantic parsing}.
\newblock In \emph{Proceedings of the 56th Annual Meeting of the Association
  for Computational Linguistics (Volume 1: Long Papers)}, pages 731--742,
  Melbourne, Australia. Association for Computational Linguistics.

\bibitem[{Dyer et~al.(2016)Dyer, Kuncoro, Ballesteros, and
  Smith}]{dyer2016rnng}
Chris Dyer, Adhiguna Kuncoro, Miguel Ballesteros, and Noah~A. Smith. 2016.
\newblock Recurrent neural network grammars.
\newblock In \emph{Proceedings of the 2016 Conference of the North {A}merican
  Chapter of the Association for Computational Linguistics: Human Language
  Technologies}, pages 199--209, San Diego, California. Association for
  Computational Linguistics.

\bibitem[{Fernandez~Astudillo et~al.(2020)Fernandez~Astudillo, Ballesteros,
  Naseem, Blodgett, and
  Florian}]{fernandez-astudillo2020transition_transformer}
Ram{\'o}n Fernandez~Astudillo, Miguel Ballesteros, Tahira Naseem, Austin
  Blodgett, and Radu Florian. 2020.
\newblock \href {https://doi.org/10.18653/v1/2020.findings-emnlp.89}
  {Transition-based parsing with stack-transformers}.
\newblock In \emph{EMNLP Findings}.

\bibitem[{Gu et~al.(2016)Gu, Lu, Li, and Li}]{gu2016copynet}
Jiatao Gu, Zhengdong Lu, Hang Li, and Victor~O.K. Li. 2016.
\newblock Incorporating copying mechanism in sequence-to-sequence learning.
\newblock In \emph{ACL}.

\bibitem[{Hwang et~al.(2019)Hwang, Kim, Yim, Seo, Park, Park, Lee, Lee, and
  Lee}]{hwang2019pot}
Wonseok Hwang, Seonghyeon Kim, Jinyeong Yim, Minjoon Seo, Seunghyun Park,
  Sungrae Park, Junyeop Lee, Bado Lee, and Hwalsuk Lee. 2019.
\newblock Post-ocr parsing: building simple and robust parser via bio tagging.
\newblock In \emph{Workshop on Document Intelligence at NeurIPS 2019}.

\bibitem[{Hwang et~al.(2020)Hwang, Yim, Park, Yang, and Seo}]{hwang2020spade}
Wonseok Hwang, Jinyeong Yim, Seunghyun Park, Sohee Yang, and Minjoon Seo. 2020.
\newblock \href {http://arxiv.org/abs/2005.00642} {Spatial dependency parsing
  for semi-structured document information extraction}.

\bibitem[{Katti et~al.(2018)Katti, Reisswig, Guder, Brarda, Bickel, H{\"o}hne,
  and Faddoul}]{katti2018chargrid}
Anoop~R Katti, Christian Reisswig, Cordula Guder, Sebastian Brarda, Steffen
  Bickel, Johannes H{\"o}hne, and Jean~Baptiste Faddoul. 2018.
\newblock {C}hargrid: Towards understanding 2{D} documents.
\newblock In \emph{EMNLP}.

\bibitem[{Kingma and Ba(2015)}]{kingam2015adam}
Diederik~P. Kingma and Jimmy Ba. 2015.
\newblock Adam: {A} method for stochastic optimization.
\newblock In \emph{ICLR}.

\bibitem[{Liu et~al.(2019)Liu, Gao, Zhang, and Zhao}]{liu2019graph}
Xiaojing Liu, Feiyu Gao, Qiong Zhang, and Huasha Zhao. 2019.
\newblock Graph convolution for multimodal information extraction from visually
  rich documents.
\newblock In \emph{NAACL}.

\bibitem[{Majumder et~al.(2020)Majumder, Potti, Tata, Wendt, Zhao, and
  Najork}]{majumder2020representation}
Bodhisattwa~Prasad Majumder, Navneet Potti, Sandeep Tata, James~Bradley Wendt,
  Qi~Zhao, and Marc Najork. 2020.
\newblock \href {https://doi.org/10.18653/v1/2020.acl-main.580} {Representation
  learning for information extraction from form-like documents}.
\newblock In \emph{Proceedings of the 58th Annual Meeting of the Association
  for Computational Linguistics}, pages 6495--6504, Online. Association for
  Computational Linguistics.

\bibitem[{Palm et~al.(2017)Palm, Winther, and Laws}]{palm2017cloundscan}
Rasmus~Berg Palm, Ole Winther, and Florian Laws. 2017.
\newblock Cloudscan - {A} configuration-free invoice analysis system using
  recurrent neural networks.
\newblock \emph{CoRR}.

\bibitem[{Qian et~al.(2019)Qian, Santus, Jin, Guo, and
  Barzilay}]{qian2019graphie}
Yujie Qian, Enrico Santus, Zhijing Jin, Jiang Guo, and Regina Barzilay. 2019.
\newblock {G}raph{IE}: A graph-based framework for information extraction.
\newblock In \emph{NAACL}.

\bibitem[{See et~al.(2017)See, Liu, and Manning}]{see2017gatedcopy}
Abigail See, Peter~J. Liu, and Christopher~D. Manning. 2017.
\newblock Get to the point: Summarization with pointer-generator networks.
\newblock In \emph{ACL}.

\bibitem[{Vaswani et~al.(2017)Vaswani, Shazeer, Parmar, Uszkoreit, Jones,
  Gomez, Kaiser, and Polosukhin}]{vaswani2017transformer}
Ashish Vaswani, Noam Shazeer, Niki Parmar, Jakob Uszkoreit, Llion Jones,
  Aidan~N Gomez, \L~ukasz Kaiser, and Illia Polosukhin. 2017.
\newblock Attention is all you need.
\newblock In I.~Guyon, U.~V. Luxburg, S.~Bengio, H.~Wallach, R.~Fergus,
  S.~Vishwanathan, and R.~Garnett, editors, \emph{NeurIPS}.

\bibitem[{Wolf et~al.(2020)Wolf, Debut, Sanh, Chaumond, Delangue, Moi, Cistac,
  Rault, Louf, Funtowicz, Davison, Shleifer, von Platen, Ma, Jernite, Plu, Xu,
  Scao, Gugger, Drame, Lhoest, and Rush}]{wolf2020huggin}
Thomas Wolf, Lysandre Debut, Victor Sanh, Julien Chaumond, Clement Delangue,
  Anthony Moi, Pierric Cistac, Tim Rault, Rémi Louf, Morgan Funtowicz, Joe
  Davison, Sam Shleifer, Patrick von Platen, Clara Ma, Yacine Jernite, Julien
  Plu, Canwen Xu, Teven~Le Scao, Sylvain Gugger, Mariama Drame, Quentin Lhoest,
  and Alexander~M. Rush. 2020.
\newblock \href {https://www.aclweb.org/anthology/2020.emnlp-demos.6}
  {Transformers: State-of-the-art natural language processing}.
\newblock In \emph{EMNLP System Demonstrations}, Online. Association for
  Computational Linguistics.

\bibitem[{Xu et~al.(2019)Xu, Li, Cui, Huang, Wei, and Zhou}]{xu2019_layoutLM}
Yiheng Xu, Minghao Li, Lei Cui, Shaohan Huang, Furu Wei, and Ming Zhou. 2019.
\newblock Layoutlm: Pre-training of text and layout for document image
  understanding.
\newblock In \emph{KDD}.

\bibitem[{Yin and Neubig(2018)}]{yin2018TRANX}
Pengcheng Yin and Graham Neubig. 2018.
\newblock \href {https://www.aclweb.org/anthology/D18-2002} {{TRANX}: A
  transition-based neural abstract syntax parser for semantic parsing and code
  generation}.
\newblock In \emph{Proceedings of the 2018 Conference on Empirical Methods in
  Natural Language Processing: System Demonstrations}, pages 7--12, Brussels,
  Belgium. Association for Computational Linguistics.

\bibitem[{{Yu} et~al.(2020){Yu}, {Lu}, {Qi}, {Gong}, and {Xiao}}]{yu2020_pick}
Wenwen {Yu}, Ning {Lu}, Xianbiao {Qi}, Ping {Gong}, and Rong {Xiao}. 2020.
\newblock {PICK: Processing Key Information Extraction from Documents using
  Improved Graph Learning-Convolutional Networks}.
\newblock In \emph{ICPR}.

\bibitem[{Zhang and Shasha(1989)}]{zhang1989ted}
K.~Zhang and D.~Shasha. 1989.
\newblock Simple fast algorithms for the editing distance between trees and
  related problems.
\newblock \emph{SIAM J. Comput.}, 18(6):1245–1262.

\bibitem[{Zhao et~al.(2019)Zhao, Wu, and Wang}]{zhao2019cutie}
Xiaohui Zhao, Zhuo Wu, and Xiaoguang Wang. 2019.
\newblock {CUTIE:} learning to understand documents with convolutional
  universal text information extractor.
\newblock \emph{CoRR}.

\bibitem[{Zhong et~al.(2020)Zhong, ShafieiBavani, and
  Jimeno-Yepes}]{Zhong2020pubtabnet}
Xu~Zhong, Elaheh ShafieiBavani, and Antonio Jimeno-Yepes. 2020.
\newblock Image-based table recognition: data, model, and evaluation.
\newblock In \emph{ECCV}.

\end{thebibliography}
\bibliographystyle{acl_natbib}

\appendix
\onecolumn
\section{Appendix}

\subsection{\vone} \label{sec: pot detail}
Here, we explain each module of \vone\ \citep{hwang2019pot}. The serializer accepts 2D text from the OCR module and converts them into a single pseudo-1D-text. To group text segments line-by-line, the segments are merged based on their height differences. When the text segments are placed on the curved line due to physical distortion of documents as often observed in receipt images, a polynomial fit is used.
The tagging model performs IOB-tagging on the pseudo-1D-text. The model is based on BERT \citep{devlinBERT2018} except that 2D coordinate embeddings are added to input vectors. The embeddings are prepared in the same way with \ours\ (Sec.~\ref{sec: model}).
The output consists of IOB-tags of multiple fields. For inter-fields grouping (e.g. \texttt{name}, \texttt{count}, and \texttt{price} under \texttt{item} in Fig.~\ref{fig_tree}a), additional IOB-tags are introduced.
The tagged text is structured into raw parses by the tag2parse module.
Finally, the raw parses are normalized using regular expressions and various domain-specific rules. For example, a unit price ``@2,000'' is converted into ``2000''; Chinese numbers in postal codes are converted into English numerals etc. 

\subsection{Evaluation methods} \label{sec: evaluation detail}

\paragraph{$F_1$ score}
To calculate $F_1$, first a group of key-value pairs from the ground truth (gt) is matched with a group from predicted parse based on their similarity in character level.
Each predicted key-value pair is counted as true positive  if there exists exactly equal gt key-value pair in the matched group. Otherwise it is counted as false positive. Unmatched key-value pairs in ground truth are counted as false negative.

\paragraph{Tree edit distance}
Although $F_1$ can show model performance for individual fields, the group matching algorithms requires non-trivial modification per domain due to structural change in parses.
Hence, we use another metric \ted\ based on tree edit distance (TED) \citep{zhang1989ted}\footnote{https://github.com/timtadh/zhang-shasha} that can be used for any documents represented as trees.
\begin{equation}
    \ted = \text{TED(\textit{gt}, \textit{pr})} / \text{TED(\textit{gt}, $\phi$)}
\end{equation}
Here, \textit{gt}, \textit{pr}, and $\phi$ stands for ground truth, predicted, and empty trees respectively.
The process is depicted in Fig.~\ref{fig_tree}b).
To account the permutation symmetry, the node in each level is sorted before the calculation using their labels and their children's.
A similar score has been recently suggested by \citep{Zhong2020pubtabnet} for a table recognition task. 

\paragraph{Human Evaluation via A/B test} \label{sec: ab}
While predefined metrics are useful for automated evaluation, their score cannot fully reflect the overall performance. Hence, we prepare accompanying human evaluation via A/B test.
In the test, the randomly selected output of \ours\ and \vone\ are presented to human subjects with corresponding document image. Then the human subjects are asked to choose one option out of three choices: A is better, B is better, or neutral. The results of two models are randomly shown as either A or B.

\subsection{1-$F_1$ of individual fields}

\begingroup
\setlength{\tabcolsep}{3.pt} 
\renewcommand{\arraystretch}{1} 
\begin{table}[h!]
\centering
\begin{threeparttable}
\captionsetup{font=small} 
  \caption{$1-F_1$ of representative fields in Japanese name card IE task.\textsc{P} and  \textsc{W} stand for \vone\ and \ours. 
  }
  \label{tbl_individual_fields_nj}
  \tiny
  \centering
  \begin{tabular}{lcccccc|ccc}
    Model & total  & address & company name & department &  personal name  & position & email & fax &tel  \\
    \midrule
    
    \textsc{P}             &  10.5 & 16.7 & 22.1  & 16.1   & 8.6&  13.6 & 4.7 & 2.6 &4.1 \\
    \textsc{W}             &  11.0 & 17.8 & 16.5  & 19.3   & 5.5&  14.0 & 6.3 & 3.8 &7.0 \\
    \textsc{W} w/ weak-label pretraining  &  8.1  & 14.2 & 12.5  & 14.9   & 3.0&  10.8 & 4.6 & 3.0 &4.9 \\

    \bottomrule
  \end{tabular}

\end{threeparttable}
\end{table}
\endgroup

\begingroup
\setlength{\tabcolsep}{3.pt} 
\renewcommand{\arraystretch}{1} 
\begin{table}[h!]
\centering
\begin{threeparttable}
\captionsetup{font=small} 
  \caption{$1-F_1$ of representative fields in Korean receipt IE task.
  }
  \label{tbl_individual_fields_rk}
  \tiny
  \centering
  \begin{tabular}{lccccc|cccccccccc}
    Model & total  
    & \makecell{item\\name} & \makecell{store\\name} & \makecell{store\\address} & \makecell{payment\\card company}
    & \makecell{item\\count}  
    & \makecell{item\\unit price} & \makecell{item\\price} 
     & \makecell{store\\tel} & \makecell{store\\business number}
     & \makecell{payment\\card number} & \makecell{payment\\confirmed number} 
    & \makecell{payment\\date} & \makecell{payment\\time}
    & \makecell{total\\price}
    \\
    \midrule
    
    \textsc{P} & 15.7 & 35.0 & 21.9 & 24.9
    & 16.2 
    & 8.1 
    & 6.9 & 4.1 
    & 15.5 & 7.3
    & 17.7 & 28.1 
    & 10.1 & 8.0
    & 6.8
    \\
    \textsc{W} & 13.0 & 33.1 & 21.3 & 18.9 
    & 18.5 
    & 3.2 
    & 5.2 & 3.4
    & 15.4 & 10.9
    & 14.6 & 17.0 
    & 7.8 & 5.0
    & 3.9
    \\
    \bottomrule
  \end{tabular}

\end{threeparttable}
\end{table}
\endgroup

\begingroup
\setlength{\tabcolsep}{3.pt} 
\renewcommand{\arraystretch}{1} 
\begin{table}[h!]
\centering
\begin{threeparttable}
\captionsetup{font=small} 
  \caption{$1-F_1$ of representative fields in Japanese receipt IE task.
  }
  \label{tbl_individual_fields_rj}
  \tiny
  \centering
  \begin{tabular}{lcccccc|cccccccccccccc}
    Model & total  
    & \makecell{item\\name} & \makecell{store\\name} & \makecell{store\\address}  & \makecell{store\\branch name}
    & \makecell{payment\\method} 
    & \makecell{item\\count}  
    & \makecell{item\\unit price} & \makecell{item\\price} 
    
    & \makecell{store\\tel} 
    & \makecell{payment\\price} & \makecell{payment\\change price} 
    & \makecell{sub total\\price} & \makecell{sub total\\tax}
    & \makecell{transaction\\date} & \makecell{transaction\\time}
    & \makecell{total\\price} & \makecell{total\\tax}
    \\
    \midrule
    \textsc{P} & 20.5
    & 24.8 & 36.3 & 30.3 & 22.2 & 24.2 
    & 3.1
    & 5.9 & 3.1 
    & 12.8
    & 4.1 & 1.4 
    & 5.7 & 6.5
    & 17.6 & 2.2
    & 12.1 & 8.1
    \\
    \textsc{W} & 25.8 
    & 40.1 & 38.6 & 38.6 & 17.1 & 27.0
    & 13.7
    & 23.6 & 14.1
    & 14.0
    & 9.8 & 2.7
    & 13.8 & 24.1
    & 18.5 & 4.4 
    & 17.3 & 17.0
    \\
    \bottomrule
  \end{tabular}

\end{threeparttable}
\end{table}
\endgroup

\begin{figure}[h]
\centering
\includegraphics[width=0.35\textwidth]{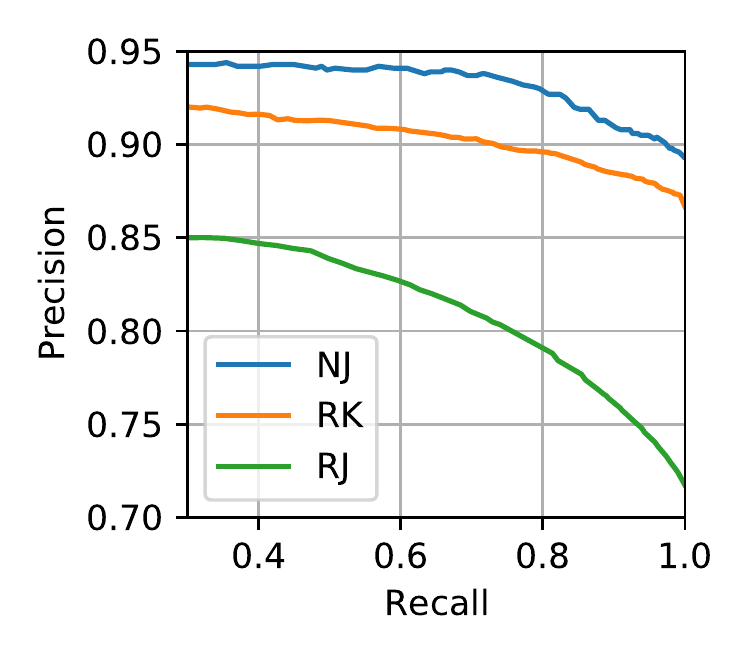}
\caption{Precision-recall curves of three IE tasks: Japanese name card (\nj), Korean receipt (\rk), and Japanese receipt (\rj). The recall rate is controlled by trimming documents of which \ours\ shows low confidence. The confidence score is calculated empirically by averaging the token generation probabilities.}
\label{fig_pr_curves}

\end{figure}

\subsection{\nj\ vs \njw} \label{sec: nj vs njw}
The \nj\ and \njw\ parses have following distinct features. (1) In \nj, same field can appear multiple times (for example, several telephone numbers can be presented in single name card) whereas in \njw, which was prepared by human before the era of deep learning, only single realization per each field is picked and saved in the database. (2) In \njw, the parse can include field that does not appear explicitly. For instance, parses can include Japanese character representation of company name even when name cards include only Chinese character representation.

\subsection{A/B test with varying layout complexity} \label{sec: ab detail}
In the name card A/B test, the examples are further separated into two groups based on the number of \texttt{address} field. First group (Lv. 1) includes  single \texttt{address} field and second group (Lv. 2) includes the multiple number of \texttt{address} field. The latter group show more diverse layout such as multicolumn and vertical text alignment. Approximately 12\% of cards belong to Lv. 2 group in the data. The result is summarized in Tbl.~\ref{tbl_ab_detail}.

\begingroup
\setlength{\tabcolsep}{5pt} 
\renewcommand{\arraystretch}{0.8} 
\begin{table}[h!]
\captionsetup{font=small} 
  \caption{A/B-test for human evaluation. 
  }
  \label{tbl_ab_detail}
  \tiny
  \centering
  \begin{tabular}{ccccccc}
      \toprule
    \multicolumn{2}{c}{} &                  
    \multicolumn{2}{c}{Training set} &                  
    \multicolumn{3}{c}{Survey results} 
    \\
    \cmidrule(r){3-4}
    \cmidrule(r){5-7}

    Task &  \# of samples &\textsc{P} &  \textsc{W} & Neutral    &\textsc{P} is better  & \textsc{W} is better  \\
    \midrule
    \nj-Lv1 &180 &\nj & \nj& 47.8 & \textbf{28.0} & 23.2 \\
    \nj-Lv2 &45 &\nj & \nj& 38.4 & 28.3 & \textbf{33.3} \\
    \midrule
    \nj-Lv1 &60 &\nj & \nj, \njw& 43.3 & 25.0 & \textbf{31.7} \\
    \nj-Lv2 &20  &\nj & \nj, \njw& 50.0 & 20.0 & \textbf{30.0} \\
    \bottomrule
  \end{tabular}

\end{table}
\endgroup

\subsection{Cost estimation}\label{sec: cost detail}

\paragraph{Annotation cost}
The annotation cost is quantified by the number of documents that can be labeled by a single annotator per hour in the name card IE task. The strong-label data requires about 5 times longer annotation time compared to the weak-label data (3rd column). 

\paragraph{Communication cost}
The tag annotators should be trained by an expert (1) to understand the connection between tagged texts and corresponding parses and (2) to become accustomed to annotation process and using tag annotation tool.
In our receipt annotation task, five tag annotators were trained by one expert for five working days. The expert needed to use one full working day. By counting 20 working days of single annotator as 1 Person Month (PM) and that of expert as 3 PM, the communication (teaching) cost is calculated as 1--2 PM \footnote{1 PM$\times 5~ \text{annotators} \times 5/20~\text{month} +$ 3 PM $\times 1~\text{expert} \times 1/20~\text{month}$ = 1.4 PM}.
On the other hand, the parse annotators just need to see the images and type human readable parses. The process similar to a summarizing documents on notes in which they are already familiar with. This minimizes the communication cost.

\paragraph{Annotation tool development cost}
The tag annotation tool should be prepared per ontology. In our own experience, the tool modification takes approximately two working days of one expert per domain (0.3 PM = 3PM $\times$ 2/20). On the other hand, the parse annotation tool does not require such modification as the format of parses are already capable of expressing arbitrary complex layout (for example JSON format can be utilized).

\paragraph{Maintenance cost}
The strong-label data can be modified only by the people trained in converting tagged text segments into parses. There is no such restriction in the weak-label data.

\paragraph{Inference cost}
We compare the inference cost of two IE systems by calculating the inference time. In name card IE task, \vone\ requires 0.4 s per document on average whereas \ours\ requires 1.4 s. In receipt IE task, \vone\ and \ours\ take 1.6 s and 2.3 s, respectively. In \vone, the serializer costs the most of inference time (Fig.~\ref{fig_v1_vs_v3task}a).
Although \ours\ takes slightly more time for the inference, the OCR module requires few seconds and the overall difference between two IE system is not significant. The time is measured on the computer equipped with Intel Xeon cpu (2.20 GHz) and P40 NVIDIA gpu with single batch.




\end{document}